\documentclass{ifacconf}

\usepackage{subcaption}
\usepackage{graphicx}      
\usepackage{natbib}        
\usepackage{amsmath}
\usepackage{amssymb}
\usepackage{booktabs}
\usepackage{multirow}
\usepackage{xcolor}
\begin{document}
\begin{frontmatter}

\title{Symmetry-Aware Steering of Equivariant Diffusion Policies: Benefits and Limits\thanksref{footnoteinfo}} 

\thanks[footnoteinfo]{
This work was supported by the National Research Foundation of Korea (NRF) grant funded by the Korea government (MSIT) (No.RS-2024-00344732). This work was also supported by the Korea Institute of Science and Technology (KIST) Institutional Program (Project No.2E33801-25-015). R. Horowitz’s research was partially supported by HKCRC.}

\author[First]{Minwoo Park\thanksref{co-first}}
\author[First]{Junwoo Chang\thanksref{co-first}}
\author[First]{Jongeun Choi}
\author[Second]{Roberto Horowitz}

\thanks[co-first]{Equally contributed.}

\address[First]{Yonsei University, Seoul 03722, Republic of Korea}
\address[Second]{University of California, Berkeley, CA 94720, USA}
\address{e-mail: \{minwoopark, junwoochang, jongeunchoi\}@yonsei.ac.kr, horowitz@berkeley.edu}

\begin{abstract}
Equivariant diffusion policies (EDPs) combine the expressivity of diffusion models with the generalization and sample efficiency afforded by geometric symmetries. While steering these policies with reinforcement learning (RL) offers a promising fine-tuning mechanism beyond demonstration data, directly applying standard (non-equivariant) RL ignores the symmetries that EDPs are designed to exploit, leading to sample inefficiency and instability. We theoretically establish that the diffusion process of an EDP is equivariant, inducing a group-invariant latent-noise MDP which is well-suited for equivariant diffusion steering. Building on this, we introduce a principled symmetry-aware steering framework and compare standard, equivariant, and approximately equivariant RL strategies across tasks with varying degrees of symmetry. While we identify the practical boundaries of strict equivariance under symmetry breaking, exploiting symmetry during the steering yields substantial benefits---enhancing sample efficiency, preventing value divergence, and achieving strong policy improvements even when EDPs are trained from extremely limited demonstrations.
\end{abstract}


\end{frontmatter}

\section{Introduction}

Leveraging geometric symmetries via group-equivariant networks improves sample efficiency and generalization in robot learning \citep{ryu2022equivariant, seo2023contact, ryu2024diffusion, seo2025equicontact, van2020mdp, wang2022so2equivariant, seo2025se}. Building on this, \textbf{Equivariant Diffusion Policies (EDPs)} \citep{wang2024equivariant} combine the expressivity of Diffusion Policies \citep{chi2023diffusionpolicy} with equivariant architectures. Despite their success, policies trained purely on demonstrations often remain suboptimal and benefit from online fine-tuning via reinforcement learning (RL) \citep{ren2024diffusion}; however, directly fine-tuning diffusion policies is computationally expensive due to backpropagation through the denoising process. \textbf{Diffusion Steering via Reinforcement Learning (DSRL)} \citep{wagenmaker2025steering} addresses this by training an RL agent to select optimal initial latent noise, treating the pre-trained policy as a black box.

Yet, the effectiveness of DSRL relies on the base policy's ability to cover the state space, which typically requires training on large-scale expert demonstrations for complex tasks. EDPs offer a compelling solution to this data dependency by generalizing efficiently from small datasets, yet steering EDPs remains unexplored.
Moreover, naively applying DSRL to EDPs ignores the symmetries exploited by the base EDP, possibly causing sample inefficiency during the steering.

In this work, we investigate symmetry-aware steering of EDPs. We establish a theoretical foundation for this by analyzing the symmetry structure of the EDP's diffusion process and the induced latent-noise MDP. Grounded on this theory, we propose a principled equivariant steering framework and investigate the relative efficacy of standard, equivariant, and approximately-equivariant RL \citep{finzi2021residual, wang2022approximately, chang2025partially}, the latter of which encourages symmetry via soft constraints. Our main contributions are as follows:
\begin{itemize}
    \item We prove that (i) the diffusion process of an EDP preserves equivariance and (ii) the induced latent-noise MDP is group-invariant, establishing when and why symmetry-preserving steering is possible.
    \item We introduce a symmetry-aware steering framework for equivariant and approximately equivariant RL agents to exploit geometric structure.
    \item Through experiments on manipulation tasks with varying symmetry degrees, including extremely limited demonstration settings, we show that symmetry-aware steering improves stability, prevents value divergence, and enhances performance, while identifying the practical boundaries of strict equivariance.
\end{itemize}

\section{Preliminaries}

\subsection{Diffusion Models}

Denoising Diffusion Probabilistic Model (DDPM) \citep{ho2020denoising} is a class of generative models that learn to approximate a data distribution $q(a_0)$ by reversing a gradual noising process.

\textbf{Forward and Reverse Process.} The forward process incrementally adds Gaussian noise to a clean sample $a_0$ over $T$ timesteps via a fixed variance schedule $\{\beta_t\}$, such that the noisy sample at step $t$ can be written as $a_t = \sqrt{\bar{\alpha}_t}a_0 + \sqrt{1-\bar{\alpha}_t}\epsilon$, $\epsilon\sim\mathcal{N}(0,I)$, where $\alpha_t = 1-\beta_t$ and $\bar{\alpha}_t=\prod_{s=1}^t\alpha_s$. The reverse process reconstructs $q(a_0)$ by iteratively denoising from a Gaussian prior $p(a_T)=\mathcal{N}(0,I)$, using a neural network $\epsilon_\theta$ trained to predict the added noise via minimizing the simplified objective $\mathcal{L} = \mathbb{E}_{t,a_0,\epsilon}[\|\epsilon - \epsilon_\theta(a_t,t)\|^2]$.

\textbf{Sampling via Langevin Dynamics.} Sampling is performed by iteratively refining the noise. The update rule at
each step can be viewed as Langevin dynamics, utilizing the learned score function $\nabla_{a_t} \log q(a_t) \propto -\epsilon_\theta(a_t, t)$ to guide samples toward high-density regions:
\begin{equation} \label{Eqn:sampling}
    a_{t-1} = \frac{1}{\sqrt{\alpha_t}} \left( a_t - \frac{1-\alpha_t}{\sqrt{1-\bar{\alpha}_t}} \epsilon_\theta(a_t, t) \right) + \sigma_t z,
\end{equation}
where $z \sim \mathcal{N}(0, I)$ represents the stochastic noise injection, and $\sigma_t$ controls the variance.

\textbf{Denoising Diffusion Implicit Models.}
For inference, we employ Denoising Diffusion Implicit Model (DDIM) \citep{song2020denoising}, whose non-Markovian formulation allows for accelerated sampling with fewer denoising steps and supports not only stochastic but also deterministic sampling ($\sigma_t=0$ in Eq.~\ref{Eqn:sampling}). Under this setting, a diffusion policy can be treated as a deterministic mapping $\pi(s, w)$ from state $s$ and latent noise $w$ to action $a$.

\subsection{Reinforcement Learning}
We consider reinforcement learning in a Markov decision process (MDP) $\mathcal{M} := (\mathcal{S}, \mathcal{A}, P, p_0, r, \gamma)$, where $\mathcal{S}$ denotes the state space, $\mathcal{A}$ the action space, $P$ the transition probability distribution, $p_0$ the initial state distribution, $r$ the reward, and $\gamma$ the discount factor. The objective is to learn a policy $\pi(a|s)$ that maximizes the expected discounted return $\mathbb{E}\big[\sum_{t=0}^{\infty} \gamma^t r(s_t,a_t)\big].$ We build on Soft Actor–Critic (SAC) \citep{haarnoja2018soft}, a widely used off-policy actor–critic method.

\subsection{Leveraging Group Symmetry}
\textbf{Group Equivariance.} A function $f: V_{\text{in}} \to V_{\text{out}}$ is equivariant to a group $G$ if it commutes with group transformations. Formally, given group representations $\rho_{\text{in}}: G \rightarrow \mathrm{GL}(V_{\text{in}})$ and $\rho_{\text{out}}: G \rightarrow \mathrm{GL}(V_{\text{out}})$, $f$ is equivariant if $f(\rho_{\text{in}}(g)x) = \rho_{\text{out}}(g) f(x)$ for all $g \in G, x \in V_{\text{in}}$. For brevity, we denote this as $f(gx) = gf(x)$. In this work, we focus on the rotation group $SO(2)$, approximated by the cyclic subgroup $C_N$.

\textbf{Group-Invariant MDPs.} A $G$-invariant MDP $\mathcal{M}_G$ \citep{wang2022so2equivariant} is an MDP abstraction that reduces the original MDP model size \citep{ravindran2001symmetries}. It satisfies two key conditions: 1) \textit{Reward Invariance}: $r(s,a)=r(gs,ga)$, and 2) \textit{Transition Invariance}: $P(s'|s,a)=P(gs'|gs,ga)$. In such MDPs, the optimal policy $\pi^*$ is $G$-equivariant ($\pi^*(gs)=g\pi^*(s)$) and the optimal critic $Q^*$ is $G$-invariant ($Q^*(gs, ga)=Q^*(s,a)$) \citep{wang2022so2equivariant}.

\section{Symmetry in Diffusion Process}

In this section, we derive that the diffusion process of an equivariant diffusion policy preserves equivariance throughout, following the formulation of \citet{ryu2024diffusion} adapted to the DDPM framework.

\subsection{Symmetry in Task} \label{Sec:formulation}
We consider a robot manipulation task with inherent symmetry defined by a group $G$. We assume $G$ is a group of transformations acting on state and action spaces, where each element $g \in G$ is a volume-preserving transformation, i.e. $|\det(J_g)|=1$, with $J_g$ denoting the Jacobian matrix of $g$. The optimal policy for this task is described by a target distribution $q(a_0|s)$, conditioned on the state $s$. We assume this distribution to be $G$-equivariant, satisfying $q(ga_0|gs) = q(a_0|s)$. Our goal is to approximate this with a diffusion model, $p_\theta(a_0|s)$.

\subsection{Equivariant Diffusion Process}
We begin by defining the property a diffusion kernel must satisfy to be compatible with the symmetry in task.

\begin{defn} \label{def:inv_ker}
A diffusion kernel $q(a_t|a_0)$ is defined as $G$-invariant if it satisfies the following condition for any group element $g \in G$:
$$ q(ga_t|ga_0) = q(a_t|a_0). $$
\end{defn}

\begin{rem} \label{rem:inv_standard_ker}
The standard diffusion kernel is a $G$-invariant kernel for rotation groups. Its probability density is a function of the squared $L_2$ norm $\|a_t - \sqrt{\bar{\alpha}_t}a_0\|^2$, which is invariant under rotation transformations.
\end{rem}

The following proposition proves that equivariance is preserved throughout the forward diffusion process, provided the process starts from an equivariant distribution.

\begin{prop} \label{prop:forward_equiv}
    Let the clean action distribution $q(a_0|s)$ be $G$-equivariant. If this distribution is perturbed by a $G$-invariant diffusion kernel $q(a_t|a_0)$, the resulting diffused marginal action distribution at step $t$,
    $$ q(a_t|s) = \int q(a_0|s) q(a_t|a_0) da_0 $$
    is also $G$-equivariant, satisfying $q(ga_t|gs) = q(a_t|s)$.
\end{prop}

\begin{pf}
    For a transformed state $gs$, the diffused marginal distribution is:
    $$ q(ga_t|gs) = \int q(a'_0|gs) q(ga_t|a'_0) da'_0. $$
    By the premise of $G$-equivariance for $q(a_0|s)$, we have $q(a'_0|gs) = q(g^{-1}a'_0|s)$. Performing a change of variables $a_0 = g^{-1}a'_0$ (where $da'_0 = da_0$ since $G$ is volume-preserving), the integral becomes:
    $$ q(ga_t|gs) = \int q(a_0|s) q(ga_t|ga_0) da_0. $$
    Since the kernel is $G$-invariant, $q(ga_t|ga_0) = q(a_t|a_0)$, this simplifies to:
    $$ q(ga_t|gs) = \int q(a_0|s) q(a_t|a_0) da_0 = q(a_t|s). $$
    Thus, the diffused marginal distribution is $G$-equivariant.
\end{pf}

Proposition~\ref{prop:forward_equiv} establishes that the forward diffusion process from a $G$-equivariant distribution results in a $G$-equivariant marginal $q(a_T|s)$.
Since effective generative modeling assumes the prior $p(a_T)$ closely approximates this marginal, we assume the initial noise distribution to be $G$-equivariant.
A standard EDP satisfies this via an isotropic Gaussian prior $\mathcal{N}(0,I)$, which is $G$-invariant for rotation groups; for a steering policy that learns a state-dependent prior, we accordingly assume $G$-equivariance.
Note that Proposition~\ref{prop:forward_equiv} serves this motivating role and is not directly invoked in the proof of Proposition~\ref{prop:GinvlatentMDP}.

\section{Symmetry-Preserving Steering with Equivariant Reinforcement Learning}


In this section, we formalize the steering problem within the latent-action MDP framework proposed by \citet{wagenmaker2025steering} and show its group-invariance. The latent-action MDP defines the environment in which the reinforcement learning agent operates to steer the pre-trained diffusion policy. We provide a theoretical justification for our method via two results: Proposition~\ref{prop:equiv_noise} extends the equivariance result of \citet{wang2024equivariant} by explicitly treating the initial noise $w = a_T$ as an input to the deterministic mapping, and Proposition~\ref{prop:GinvlatentMDP} establishes that the induced latent-action MDP is $G$-invariant, justifying the use of equivariant RL for steering.

\subsection{Theory of Equivariant Diffusion Steering}

Consistent with the task symmetry defined in Section~\ref{Sec:formulation}, we define the underlying environment $\mathcal{M}$ as a $G$-invariant MDP $\mathcal{M}_G$ under the symmetry group $G$. We use a DDPM \citep{ho2020denoising}-based policy during training but adopt the DDIM \citep{song2020denoising} formulation for inference. For simplicity, we assume deterministic sampling ($\sigma_t = 0$ in Eq.~\ref{Eqn:sampling}), reducing the denoising process to a deterministic mapping; note that the equivariance results hold in expectation even under stochastic sampling, since the injected noise $z \sim \mathcal{N}(0, I)$ under stochastic sampling is rotation-invariant. Given a pre-trained equivariant diffusion policy $\pi_\mathrm{edp}$, we define $\pi_\mathrm{edp}^\mathcal{W}: \mathcal{S} \times \mathcal{W} \to \mathcal{A}$ as the deterministic mapping which maps a state $s\in\mathcal{S}$ and a latent noise $w \in \mathcal{W}$ into an action $a\in\mathcal{A}$. We first state the equivariance with respect to the latent noise.

\begin{prop} \label{prop:equiv_noise}
    The deterministic mapping $\pi_\mathrm{edp}^\mathcal{W}$ induced by an equivariant diffusion policy $\pi_\mathrm{edp}$ is equivariant with respect to the state and the initial latent noise, i.e., $\pi_\mathrm{edp}^\mathcal{W}(gs, gw) = g\pi_\mathrm{edp}^\mathcal{W}(s, w)$.
\end{prop}

\begin{pf}
    Let $\pi_\mathrm{edp}^\mathcal{W}(s, w)$ denote the full denoising process of the EDP, which maps a state $s$ and an initial noise $w=a_T$ to a final action. This process is defined by iteratively applying a one-step denoising function $f_t$ for timesteps $t=T, \dots, 1$. Each step is defined as:
    $$ a_{t-1} = f_t(s, a_t) = \alpha_t(a_t - \gamma_t \epsilon_\theta(s, a_t, t)) $$
    where $\epsilon_\theta$ is the noise prediction network and $\alpha_t, \gamma_t$ are coefficients from the noise schedule.
    
    The network $\epsilon_\theta$ is equivariant by construction:
    $$ \epsilon_\theta(gs, ga_t, t) = g\epsilon_\theta(s, a_t, t) $$
    Consequently, the one-step function $f_t$ is also equivariant:
    \begin{align*}
        f_t(gs, ga_t) &= \alpha_t(ga_t - \gamma_t \epsilon_\theta(gs, ga_t, t)) \\
        &= \alpha_t(ga_t - \gamma_t g\epsilon_\theta(s, a_t, t)) \\
        &= g \cdot \alpha_t(a_t - \gamma_t \epsilon_\theta(s, a_t, t)) \\
        &= g \cdot f_t(s, a_t)
    \end{align*}
    The full denoising process is a composition of these equivariant one-step functions:
    $$ \pi_\mathrm{edp}^\mathcal{W}(s, w) = (f_1(s,\cdot) \circ f_2(s,\cdot) \circ \dots \circ f_T(s,\cdot))(w) $$
    Since a composition of equivariant functions is itself equivariant, the entire policy $\pi_\mathrm{edp}^\mathcal{W}$ is equivariant with respect to the state and initial noise, which concludes the proof:
    $$ \pi_\mathrm{edp}^\mathcal{W}(gs, gw) = g\pi_\mathrm{edp}^\mathcal{W}(s, w) $$
\end{pf}

\begin{figure}
\begin{center}
\includegraphics[width=0.40\textwidth]{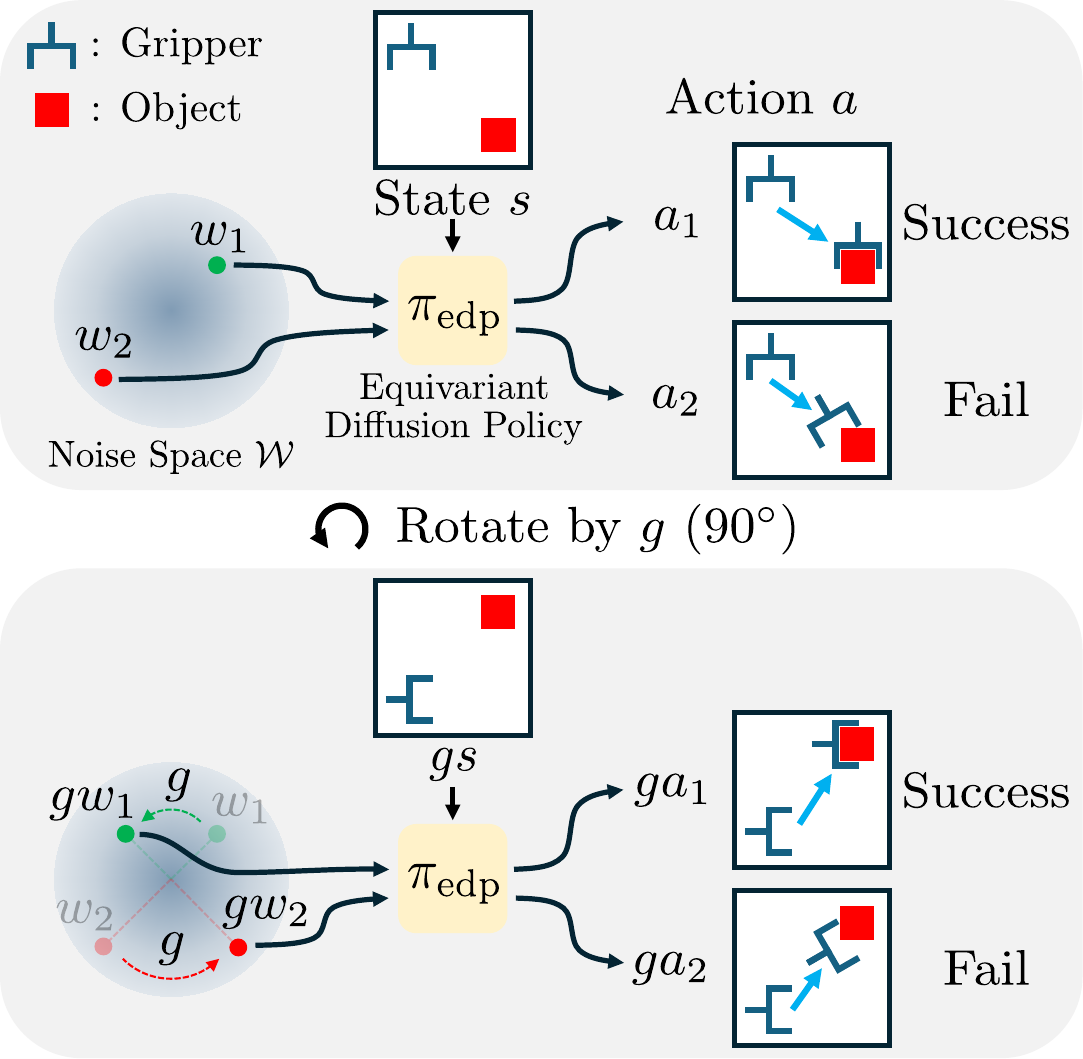}    
\caption{\textbf{Equivariance in the latent-noise space.} An EDP $\pi_\mathrm{edp}$ is equivariant with respect to state $s$ and initial latent noise $w$. \textbf{Top}: For a grasping task, optimal latent noise $w_1$ generates an optimal action $a_1$, while $w_2$ leads to suboptimal action $a_2$. \textbf{Bottom}: Under a $90^\circ$ rotation $g$, an equivariant steering policy trained to select $w_1$ for $s$ is guaranteed to select $gw_1$ for $gs$, generating the optimal action $ga_1$.}
\label{fig:latent_equiv}
\end{center}
\end{figure}

The visualization of equivariance in the latent-noise space is presented in Figure~\ref{fig:latent_equiv}. Using Proposition~\ref{prop:equiv_noise}, we now show that the latent-action MDP of an equivariant diffusion policy is $G$-invariant. With the mapping $\pi_\mathrm{edp}^\mathcal{W}$, we can transform the action space $\mathcal{A}$ of the MDP $\mathcal{M}$ to a latent-noise space $\mathcal{W}$, thereby forming a latent-action MDP
\[
\mathcal{M}^\mathcal{W}:=(\mathcal{S},\mathcal{W},P^\mathcal{W},p_0,r^\mathcal{W},\gamma),
\]
where
\[
P^\mathcal{W}(\cdot|s,w):=P(\cdot|s,\pi_\mathrm{edp}^\mathcal{W}(s,w)),
\]
\[
r^\mathcal{W}(s,w):=r(s,\pi_\mathrm{edp}^\mathcal{W}(s,w)).
\]

\begin{prop} \label{prop:GinvlatentMDP}
    The latent-action MDP, $\mathcal{M}^\mathcal{W}$, which is derived from a $G$-invariant MDP $\mathcal{M}_G$ and an equivariant mapping $\pi_\mathrm{edp}^\mathcal{W}$, is also a $G$-invariant MDP.
\end{prop}

\begin{pf}
    We show that $\mathcal{M}^\mathcal{W}$ satisfies both conditions of $G$-invariance.
    
    \textit{1. Reward Invariance.}
    By the definition of $r^\mathcal{W}$,
    \[
    r^\mathcal{W}(gs, gw) = r(gs, \pi_\mathrm{edp}^\mathcal{W}(gs, gw)).
    \]
    Applying Proposition~\ref{prop:equiv_noise} and the $G$-invariance of $r$ to the right-hand side yields:
    \[
    r(gs, g\pi_\mathrm{edp}^\mathcal{W}(s, w))
    = r(s, \pi_\mathrm{edp}^\mathcal{W}(s, w)),
    \]
    which equals $r^\mathcal{W}(s, w)$ by definition.
    Thus,
    \[
    r^\mathcal{W}(gs, gw) = r^\mathcal{W}(s, w).
    \]
    
    \textit{2. Transition Invariance.}
    By the definition of $P^\mathcal{W}$,
    \[
    P^\mathcal{W}(gs' \mid gs, gw)
    = P(gs' \mid gs, \pi_\mathrm{edp}^\mathcal{W}(gs, gw)).
    \]
    Applying Proposition~\ref{prop:equiv_noise} and the $G$-invariance of $P$ to the right-hand side yields:
    \[
    P(gs' \mid gs, g\pi_\mathrm{edp}^\mathcal{W}(s, w))
    = P(s' \mid s, \pi_\mathrm{edp}^\mathcal{W}(s, w)).
    \]
    Therefore,
    \[
    P^\mathcal{W}(gs' \mid gs, gw)
    = P^\mathcal{W}(s' \mid s, w).
    \]
    
    Since both conditions are met, the latent-action MDP $\mathcal{M}^\mathcal{W}$ is $G$-invariant.
\end{pf}

Having established that the latent-action MDP is $G$-invariant, we now discuss the optimal strategy for steering.

\subsection{Equivariant Diffusion Steering with Equivariant SAC}
Given that the latent-action MDP $\mathcal{M}^\mathcal{W}$ is $G$-invariant, we use equivariant Soft Actor-Critic (SAC) \citep{haarnoja2018soft} to maximize the expected return. Specifically, we train a $G$-equivariant actor to learn the optimal steering policy, jointly with $G$-invariant critics that approximate the optimal action-value (Q) function. As established by \citet{wang2022so2equivariant}, in a group-invariant MDP, the optimal $Q$-function is inherently $G$-invariant and the optimal policy is $G$-equivariant. This theoretical property justifies our architectural choice to enforce these geometric constraints, ensuring the steering agent efficiently learns the optimal policy while respecting the underlying symmetry.

\section{Simulation Experiments}

\begin{figure}
    \centering
    \begin{subfigure}[b]{0.25\columnwidth}
        \includegraphics[width=\textwidth]{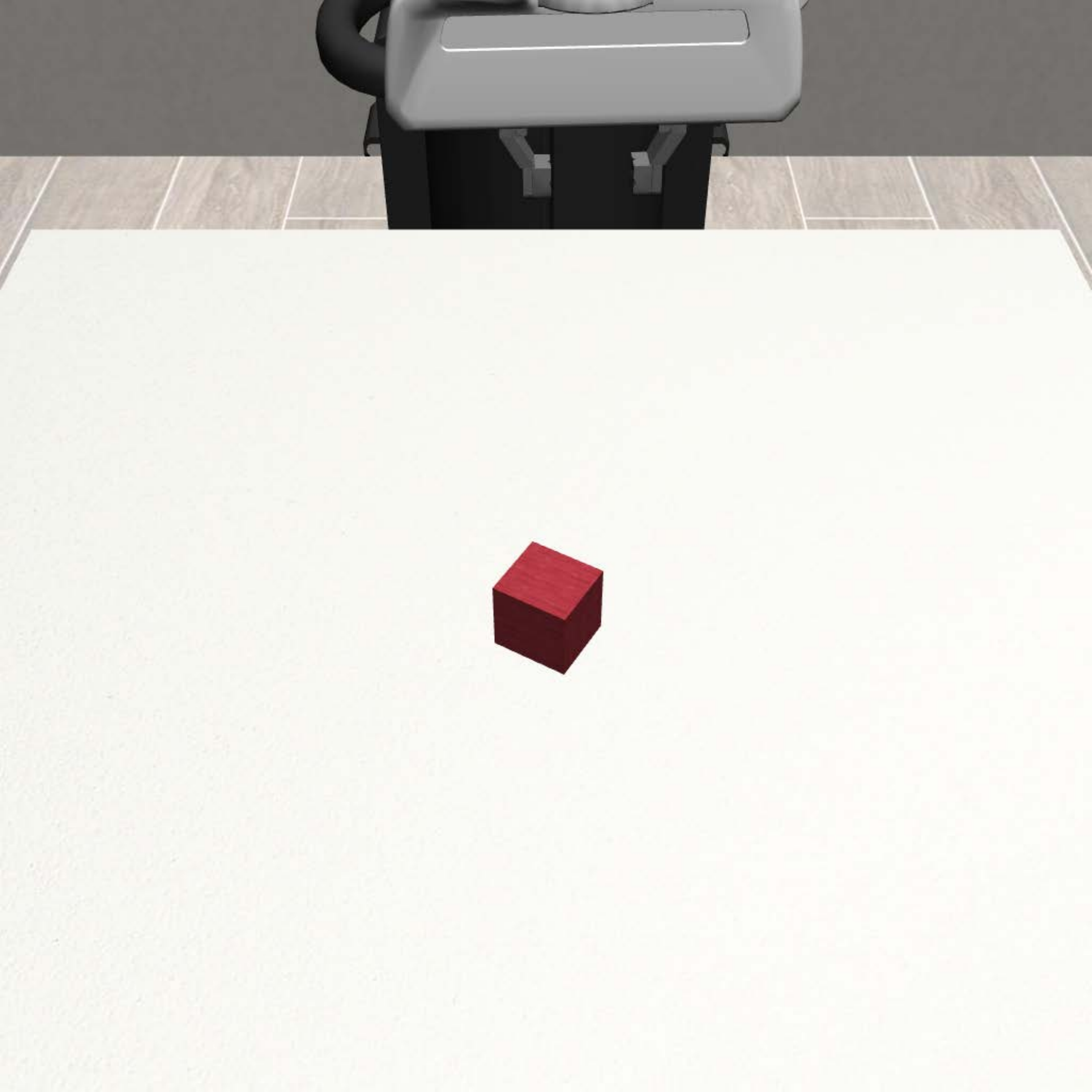}
        \caption{Lift}
    \end{subfigure}
    \hspace{0.05\columnwidth}
    \begin{subfigure}[b]{0.25\columnwidth}
        \includegraphics[width=\textwidth]{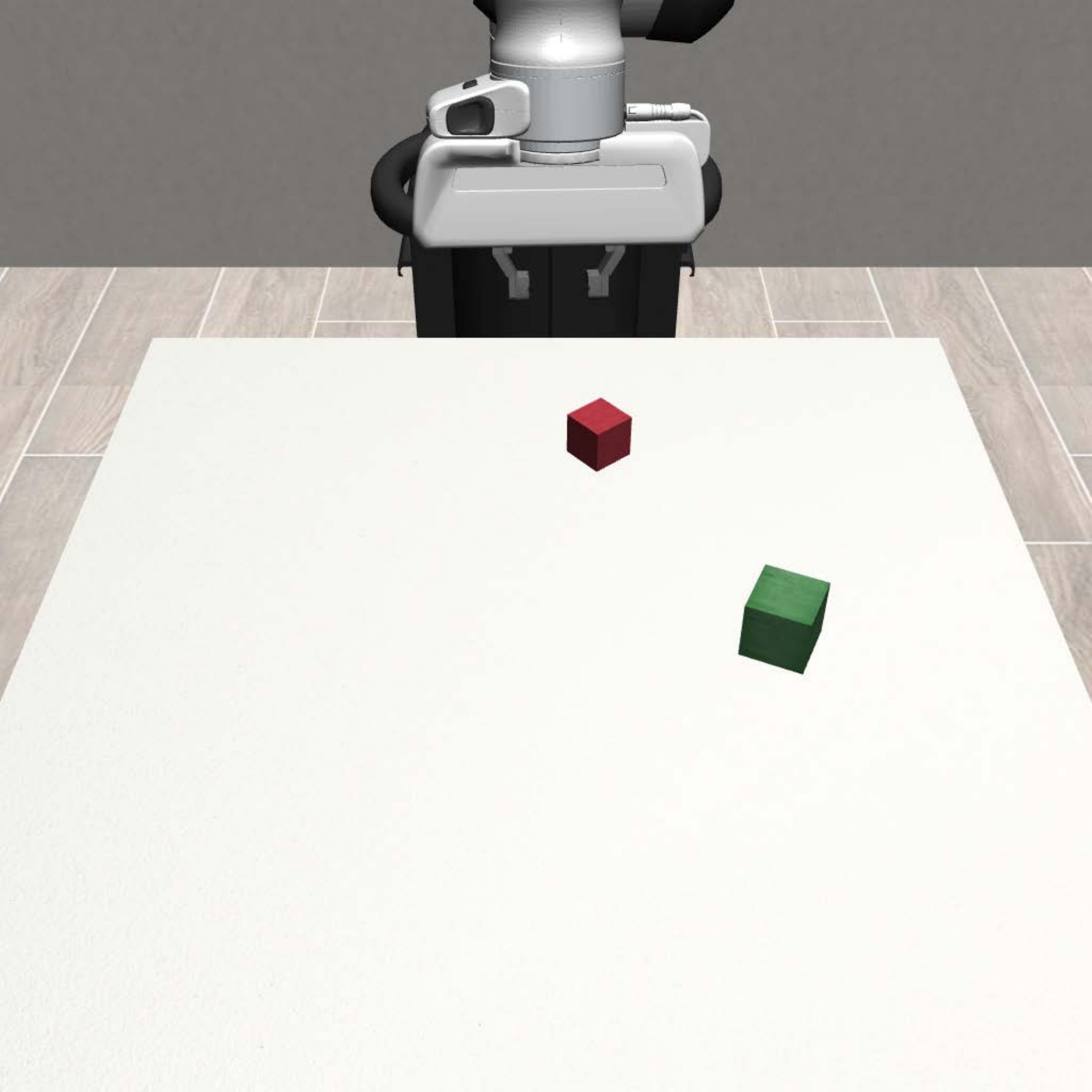}
        \caption{Stack D1}
    \end{subfigure}
    \hspace{0.05\columnwidth}
    \begin{subfigure}[b]{0.25\columnwidth}
        \includegraphics[width=\textwidth]{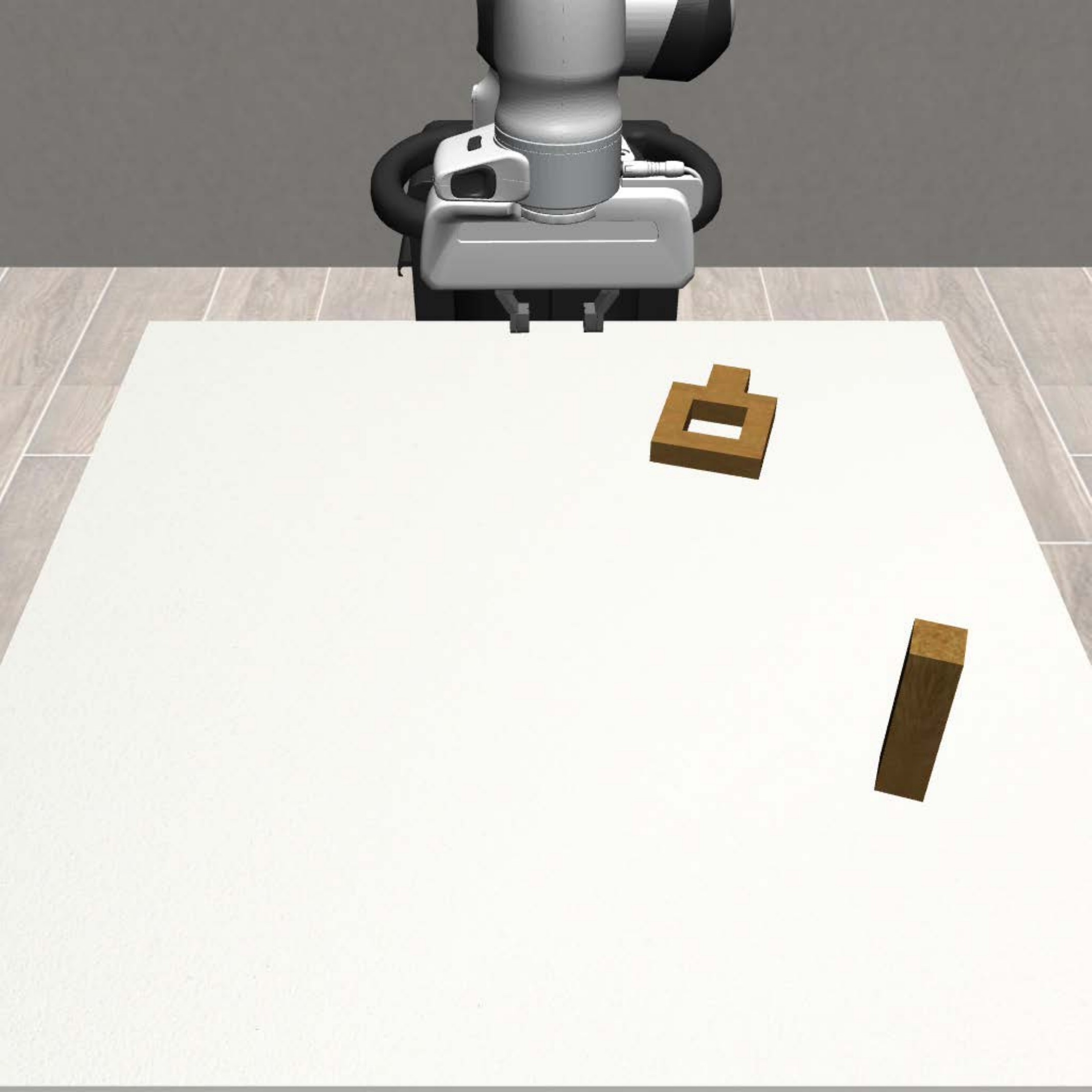}
        \caption{Square D2}
    \end{subfigure}
    \caption{\textbf{Task visualizations.} \textbf{(a) Lift}: Lifting a red cube. \textbf{(b) Stack D1}: Stacking a red cube onto a green cube. \textbf{(c) Square D2}: Inserting a square nut into a peg. Initial poses are randomized per episode.}
    \label{fig:task}
\end{figure}

We conduct experiments on tasks from \textbf{Robomimic} \citep{mandlekar2021matters} and \textbf{MimicGen} \citep{mandlekar2023mimicgen} (Figure~\ref{fig:task}). All tasks use a sparse binary reward (0 for success, $-1$ for failure) at each environment step.
Our experiment consists of two stages: pre-training state- or image-based EDPs \citep{wang2024equivariant} on demonstrations, then steering via RL in the latent-noise space \citep{wagenmaker2025steering}. We mainly compare three steering methods:
\begin{itemize}
    \item \textbf{(Standard) DSRL}: The baseline steering approach using a non-equivariant policy and critic, as proposed by \citet{wagenmaker2025steering}.
    \item \textbf{Equi-DSRL}: Our proposed method, employing a $G$-equivariant steering policy and $G$-invariant critics designed to follow the system's underlying symmetries.
    \item \textbf{Approx-Equi-DSRL}: A variant using approximately $G$-equivariant steering policy and approximately $G$-invariant critics \citep{finzi2021residual}.
\end{itemize}

\begin{figure}
\begin{center}
\includegraphics[width=0.48\textwidth]{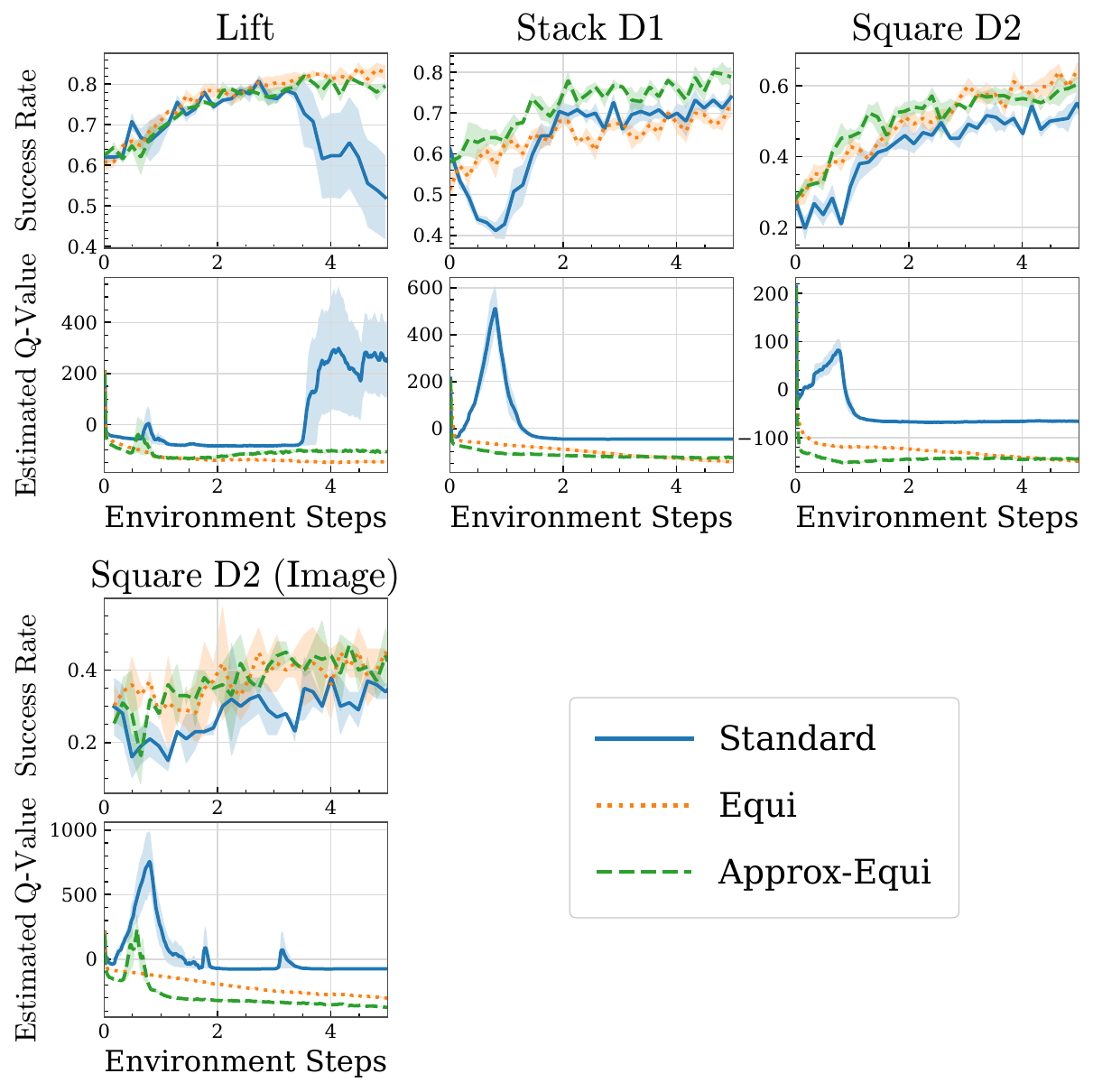}
\caption{\textbf{Performance comparison.} Results are averaged over 5M environment steps, using five random seeds (two for image-based). Shaded regions denote standard error.} 
\label{fig:performance_comparison}
\end{center}
\end{figure}

We report the mean success rate and standard error over five random seeds (three for 
image-based experiments), with periodic evaluations of 50 episodes each. Training stability is assessed via mean $Q$-values of sampled transitions. Results are shown in Figure~\ref{fig:performance_comparison} and Table~\ref{tb:results}.

\subsection{Implementation Details} \label{sec:impl}

\textbf{Base Equivariant Diffusion Policy.} We adapt the $SO(2)$-equivariant framework of \citet{wang2024equivariant} to a state-based setting. The observation $s$ includes proprioceptive and object states, while the action $a$ consists of absolute gripper poses and widths for the next 4 timesteps (see Appendix~\ref{Appendix:rep} for representation details). We employ a $C_8$-equivariant MLP backbone with \texttt{escnn} linear layers \citep{cesa2022a}. Training uses the DDPM objective \citep{ho2020denoising} (100 steps), while inference employs deterministic DDIM sampling \citep{song2020denoising} (5 steps).

\textbf{Steering Algorithm.} We utilize the DSRL-SAC algorithm \citep{wagenmaker2025steering} operating in the latent-noise space $\mathcal{W}$.
All actor and critic networks use 3-layer MLPs with matched parameter counts (DSRL: 2048 units; Equi-DSRL: 768 units with $C_8$ repr; Approx-Equi-DSRL: parallel equivariant/non-equivariant layers of 256 units each with weight decays $10^{-5}$/$10^{-2}$ \citep{finzi2021residual}). Other hyperparameters follow \citet{wagenmaker2025steering}.

\subsection{Steering in the Low-Data Regime}

\textbf{Experimental Setup.}
We use the Lift task of Robomimic with an EDP pre-trained on only three demonstrations, creating a testbed with low-capability base policy.

\textbf{Results.} Both Equi-DSRL and Approx-Equi-DSRL prove highly effective, with Equi-DSRL achieving the highest peak success rate. In contrast, standard DSRL exhibited severe instability; despite showing initial improvement, it frequently (3 out of 5 seeds) suffered from performance collapse attributed to $Q$-value divergence.

\subsection{Steering for Complex Tasks}

\textbf{Experimental Setup.} We evaluate performance on more complex manipulation tasks, Stack D1 and Square D2 from MimicGen, which involve a broader initial state distribution and require greater precision. The base EDPs are pre-trained on 100 and 200 demonstrations, respectively.

\textbf{Results.} Approx-Equi-DSRL demonstrated the most robust performance overall. In Stack D1, Equi-DSRL exhibited the lowest performance, while in Square D2 both equivariant methods performed comparably. Standard DSRL suffered from early $Q$-value divergence across both tasks and, although it eventually recovered, remained the lowest-performing method in Square D2 and failed to match the best strategy in Stack D1.

\subsection{Steering with Visual Observations}

\textbf{Experimental Setup.} We evaluate on Square D2 with image-based inputs to assess generalization beyond state-based observations. The base policy is an image-based EDP with an equivariant CNN encoder ($C_8$, 76$\times$76 crop, 128 regular representation channels). During steering, the encoder is frozen and its output feature vector serves as input to the SAC actor and critics in place of the state observation. All other settings follow 
Section~\ref{sec:impl}.

\textbf{Results.} Standard DSRL showed marginal improvement, while both Equi-DSRL and Approx-Equi-DSRL achieved comparable performance gains, demonstrating that symmetry-aware steering remains effective under image-based observations.

\begin{table}
\begin{center}
\caption{Maximum evaluation success rates.}
\label{tb:results}
\begin{tabular}{c|c|c|c|c}
\toprule
\multirow{2}{*}{Task} & \textbf{Before} & \multirow{2}{*}{\textbf{DSRL}} & \textbf{Equi-} & \textbf{Approx-Equi-} \\
 & \textbf{Steering} & & \textbf{DSRL} & \textbf{DSRL} \\
\midrule
Lift      & 0.617 & 0.808 & \textbf{0.840} & 0.820 \\
Stack D1  & 0.568 & 0.740 & 0.728 & \textbf{0.800} \\
Square D2 & 0.275 & 0.552 & \textbf{0.644} & 0.604 \\
\cmidrule{1-5}
\begin{tabular}[c]{@{}c@{}} Square D2 \\ (Image) \end{tabular} & 0.283 & 0.380 & 0.450 & \textbf{0.470} \\
\bottomrule
\end{tabular}
\end{center}
\end{table}

\subsection{Discussion}
We discuss three key aspects of our results.

\textbf{Task-Dependent Benefits of Equivariance.} The sample efficiency gains from equivariant steering are more pronounced in complex tasks, where broader workspace utilization and greater rotational variation make symmetry exploitation more valuable. In Lift, an object is confined to a small range, limiting the benefit of rotational symmetry. In contrast, Stack D1 and Square D2 involve wider $xy$ initialization and more complex manipulation, corresponding to a higher level of equivariance \citep{wang2024equivariant}.

\textbf{Impact of Symmetry Breaking.} In Stack D1, Equi-DSRL underperformed the standard DSRL, despite its theoretical efficiency. We attribute this to symmetry breaking inherent in realistic robotic manipulation, specifically caused by joint limits and kinematic singularities. The strict symmetry enforcement in Equi-DSRL could become a liability in such settings, underscoring the advantage of Approx-Equi-DSRL, which relaxes these constraints to adapt to environmental asymmetries.

\textbf{Preventing Instability from High Update Ratios.} While high update-to-data (UTD) ratios are recommended for DSRL \citep{wagenmaker2025steering}, such regimes are prone to $Q$-value divergence in off-policy actor-critic methods, as observed in standard DSRL across all tasks. We attribute this to the overestimation of out-of-distribution (OOD) actions \citep{hussing2025dissecting}, which we believe can be effectively mitigated in equivariant DSRL. By enforcing the constraint $Q(s, a) = Q(gs, ga)\;\forall g\in G$, the group-invariant critic treats a single transition as a learning signal for all symmetrically equivalent states. This structural prior reduces the size of the solution space \citep{van2020mdp}, implicitly shrinking the set of OOD actions. This suppresses $Q$-value overestimation and prevents divergence under high-UTD learning.
We validate value estimation accuracy by comparing estimated $Q$-values against Monte Carlo (MC) returns averaged over 50 evaluation episodes on Square D2 after 5M environment steps. Standard DSRL yields a mean bias of $69.8$, confirming significant overestimation. Equi-DSRL and Approx-Equi-DSRL yield $-17.0$ and $-9.65$, respectively---slightly negative, indicating mild underestimation, but substantially more accurate than standard DSRL.

\section{Conclusion and Future Work}

We propose a symmetry-aware steering framework for EDPs, providing a theory for equivariant diffusion steering and demonstrating substantial improvements even from limited demonstrations.
Future work includes extending to richer symmetry groups such as SE(3), real-robot validation, and comparison with alternative fine-tuning approaches such as DPPO \citep{ren2024diffusion}.

\section*{DECLARATION OF GENERATIVE AI AND AI-ASSISTED TECHNOLOGIES IN THE WRITING PROCESS}

During the preparation of this work, the authors used Gemini and Claude in order to generate code snippets and polish texts. After using this tool, the authors reviewed and edited the content as needed and take full responsibility for the content of the publication.


\bibliography{ifacconf}             
                                                   







\appendix

\section{State and action space representations of the tasks} \label{Appendix:rep}

We use two kinds of group representations: (i) trivial representation $\rho_0$ which acts on an invariant scalar $x\in\mathbb{R}$ by $\rho_0(g)x=x$, (ii) irreducible representation (Irrep) $\rho_w$ of the group $SO(2)$ or $C_N$ which acts on a vector $x\in\mathbb{R}^2$ by a rotation of frequency $w$, $\rho_w(g)x=\begin{bmatrix} \cos{wg} & -\sin{wg} \\ \sin{wg} & \cos{wg} \end{bmatrix}x$.


Details of the representations are specified in Table ~\ref{tb:representations}. The symbol $\oplus$ denotes direct sum, and the notation $\rho^n$ indicates the $n$-fold direct sum of the representation $\rho$.


\begin{table}[h]
\caption{State and action representations.}
\label{tb:representations}
\setlength{\tabcolsep}{3pt}
\centering
\begin{tabular}{llcc}
\toprule
& Name & Dim & Rep \\
\midrule
\multirow{2}{*}{Common}
& gripper position \& rotation  & 9 & $\rho_1\oplus\rho_0\oplus\rho_1^3$ \\
& gripper joint position      & 2 & $\rho_0^2$ \\
\midrule
\multirow{2}{*}{Lift}
& cube position \& rotation           & 9 & $\rho_1\oplus\rho_0\oplus\rho_1^3$ \\
& gripper to cube position $(x,y)$    & 2 & $\rho_1$ \\
\midrule
\multirow{2}{*}{Stack D1}
& cube position \& rotation           & 9 & $\rho_1\oplus\rho_0\oplus\rho_1^3$ \\
& gripper to cube position $(x,y,z)$  & 3 & $\rho_1\oplus\rho_0$ \\
\midrule
\multirow{3}{*}{Square D2}
& nut position \& rotation            & 9 & $\rho_1\oplus\rho_0\oplus\rho_1^3$ \\
& gripper to nut position \& rotation & 9 & $\rho_0^3\oplus\rho_0^6$ \\
& peg position $(x,y,z)$              & 3 & $\rho_1\oplus\rho_0$ \\
\midrule
\multirow{2}{*}{Action}
& gripper position \& rotation  & 9 & $\rho_1\oplus\rho_0\oplus\rho_1^3$ \\
& gripper width               & 1 & $\rho_0$ \\
\bottomrule
\end{tabular}
\end{table}

\end{document}